\documentclass[sigconf, screen]{acmart}
\AtBeginDocument{%
  }
\usepackage{bm}
\usepackage{multirow}
\usepackage{bbding}
\usepackage{pifont}
\newcommand{\etal}{\textit{et al}.}
\newcommand{\ie}{\textit{i}.\textit{e}.}

\acmSubmissionID{xxxx}

\begin{document}

\title{Motion-Aware Adaptive Pixel Pruning for\\ Efficient Local Motion Deblurring}

\author{Wei Shang}
 \affiliation{
   \institution{Harbin Institute of Technology \\ \& City University of Hong Kong}
   \country{}
 }

\author{Dongwei Ren}
\authornote{Corresponding author: rendongweihit@gmail.com}
\affiliation{
	\institution{School of Computer Science and Technology, Harbin Institute of Technology}
	\country{}
}

\author{Wanying Zhang}
\affiliation{
	\institution{School of Computer Science and Technology, Harbin Institute of Technology}
	\country{}
}

\author{Pengfei Zhu}
\affiliation{
	\institution{Tianjin University \& Low-Altitude Intelligence Lab, Xiong'an National Innovation Center Technology Co., Ltd}
	\country{}
}

\author{Qinghua Hu}
\affiliation{
	\institution{College of Intelligence and Computing, Tianjin University}
	\country{}
}

\author{Wangmeng Zuo}
\affiliation{
	\institution{School of Computer Science and Technology, Harbin Institute of Technology}
	\country{}
}


\begin{abstract}
Local motion blur in digital images originates from the relative motion between dynamic objects and static imaging systems during exposure. Existing deblurring methods face significant challenges in addressing this problem due to their inefficient allocation of computational resources and inadequate handling of spatially varying blur patterns. To overcome these limitations, we first propose a trainable mask predictor that identifies blurred regions in the image. During training, we employ blur masks to exclude sharp regions. For inference optimization, we implement structural reparameterization by converting $3\times 3$ convolutions to computationally efficient $1\times 1$ convolutions, enabling pixel-level pruning of sharp areas to reduce computation. Second, we develop an intra-frame motion analyzer that translates relative pixel displacements into motion trajectories, establishing adaptive guidance for region-specific blur restoration. Our method is trained end-to-end using a combination of reconstruction loss, reblur loss, and mask loss guided by annotated blur masks. Extensive experiments demonstrate superior performance over state-of-the-art methods on both local and global blur datasets while reducing FLOPs by 49\% compared to SOTA models (e.g., LMD-ViT). The source code is available at {\url{https://github.com/shangwei5/M2AENet}}.

\end{abstract}




\maketitle

\section{Introduction}
\label{sec:intro}
Image deblurring, a fundamental task in image restoration, aims to recover a sharp image from a blurred image, which is often the result of camera shake, object motion, or a combination of both~\cite{levin2006blind} during image acquisition. The problem of image deblurring has been extensively studied, and a variety of methods have been proposed to address this challenge~\cite{fergus2006removing,shan2008high,ren2021deblurring,tsai2022stripformer,shang2025aggregating}.
In the realm of image deblurring, motion blur can be categorized into two distinct types: global motion blur and local motion blur.
In recent years, the rapid development of deep learning has witnessed the proposal of numerous image restoration methods based on Convolutional Neural Network (CNN)~\cite{pan2016blind,tao2018scale,ren2020neural,cho2021rethinking,chen2022simple} and Transformer~\cite{zamir2022restormer,wang2022uformer,kong2023efficient,mao2024loformer}, which have achieved remarkable results in image deblurring tasks. These methods are primarily designed to handle global motion blur, where the blur effect is uniform across the entire image. However, local motion blur, which is confined to specific regions of an image, presents a more formidable challenge~\cite{levin2006blind,askari2017local}.

When global deblurring methods are applied to local motion blur, several issues arise. Firstly, these methods fail to concentrate on the blurred regions~\cite{li2023real} and lead to distortion in the sharp regions of the image, potentially degrading the visual quality of the non-blurred areas~\cite{li2024adaptive}. 
Secondly, processing all regions may lead to unnecessary computational expenditure~\cite{li2024adaptive}, which becomes particularly significant when dealing with high-resolution images captured in real-world scenarios.
To address the challenge of local motion blur, early traditional restoration methods~\cite{wang1993layered,levin2006blind,askari2017local} have employed an image segmentation-based approach, which involves segmenting the image into distinct layers and processing each layer separately. 
Recently, LBAG~\cite{li2023real} represents an advancement in this direction, which is designed to enhance the capacity of CNN to focus on the blurred regions within an image. 
Building upon the success of CNN-based methods, LMD-ViT~\cite{li2024adaptive} takes a further step by integrating the capabilities of Transformer to further enhance the performance and employs an adaptive window pruning strategy to reduce unnecessary computation.
Mao~\etal~\cite{mao2024adarevd} designed a premature exit mechanism based on the difficulty of restoration for each image patch to reduce computation. 

Despite the progress made by these methods, current local deblurring techniques primarily allocate computational resources based on image patches or windows to reduce computation~\cite{li2024adaptive,mao2024adarevd}. When patches or windows contain both blurred and sharp regions, it leads to suboptimal resource allocation. Additionally, while transformer-based methods exhibit superior performance, they are generally slower in execution speed. 
In addition, although these methods reduce the computation by performing operations such as early exit or window pruning based on the blur mask, they have not solved the problem at the root cause of the motion blur.

\begin{figure} 
	\centering
	\begin{tabular}{c}
		\includegraphics[width=\linewidth]{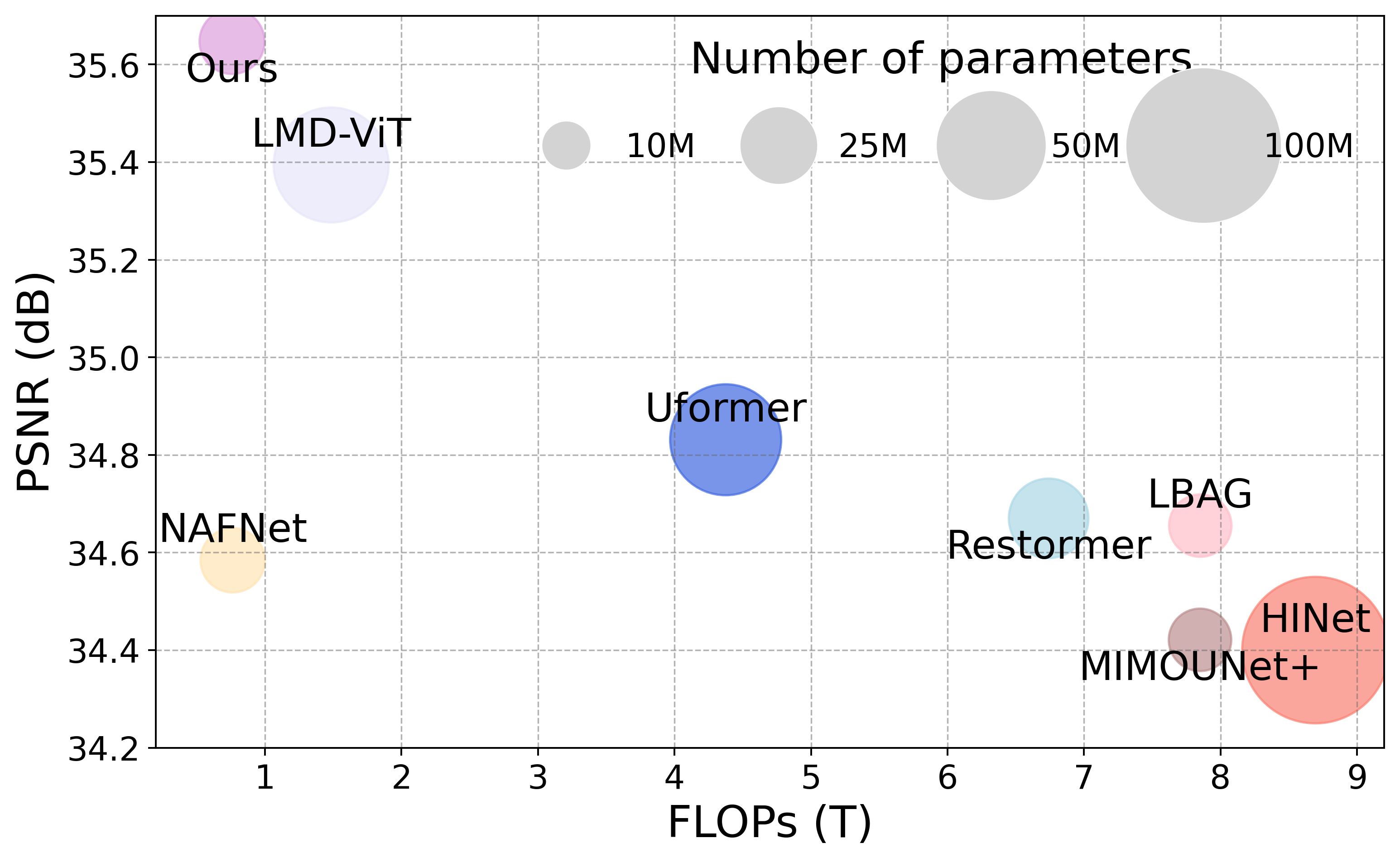}\\
	\end{tabular}
	\caption{Comparison with recent deblurring models on ReLoBlur~\cite{li2023real}. We show three metrics: deblurring performance (PSNR), model complexity (FLOPs) and the number of parameters. Our approach achieves comparable performance to state-of-the-art methods with minimal complexity and fewer parameters.
	}
	\label{fig:intro_comp}
\end{figure}
To address these issues, we propose a \textbf{M}ask- and \textbf{M}otion-\textbf{A}ware \textbf{E}fficient \textbf{Net}work (M$^2$AENet) that performs pixel-level fine-grained processing for sharp and blurred regions. By analyzing the motion trajectories caused by signal accumulation during exposure, our method identifies the pixels contributing to blur and decouples the underlying sharp pixels. 
Our method is built upon the simple yet effective lightweight network NAFNet~\cite{chen2022simple}.
We introduce a mask predictor and intra-motion analyzer at each scale to obtain a blur mask representing blurred regions and a map of pixel displacements during exposure.
To achieve pixel-level fine-grained processing for sharp and blurred regions, we propose a mask-aware adaptive pixel pruning strategy. During training, we perform element-wise multiplication between the features and the mask to exclude sharp regions while maintaining gradient backpropagation. During testing, we reparameterize convolution layers with kernels larger than 1 into $1\times 1$ convolution layers and selectively process pixels in blurred regions based on the mask to further reduce computation, then place them back into the original feature map according to the mask. To incorporate motion blur information, we analyze the motion trajectories of objects using the obtained the pixel displacements during exposure time. We compute relative displacements for each pixel, which are fed into deformable convolution to perform motion-aware deblurring explicitly.
Our entire method is end-to-end, without the need to introduce additional complex modules or pre-trained models.

In summary, our main technical contributions include
\begin{itemize}
	\item The first local motion deblurring method to dynamically distribute computational resources at the pixel level, which employs a joint optimization strategy to simultaneously constrain blur masks, intra-motion maps, and deblurring results, avoiding the need for additional complex modules or pre-trained models,
	\item The novel motion-aware adaptive pixel pruning strategy, which uses reparameterization techniques and mask-based feature sampling to reduce computational load during inference. Additionally, a motion-trajectory-based deformable convolution is proposed to perform motion-aware deblurring explicitly, 
	\item A comprehensive experimental demonstration, that M$^2$AENet significantly surpasses competing methods in terms of restoration quality on both local blur dataset~\cite{li2023real} and global blur dataset~\cite{nah2017deep} with lower model complexity and inference time.
\end{itemize}

\begin{figure*} 
	\centering
	\begin{tabular}{c}
		\includegraphics[width=0.75\linewidth]{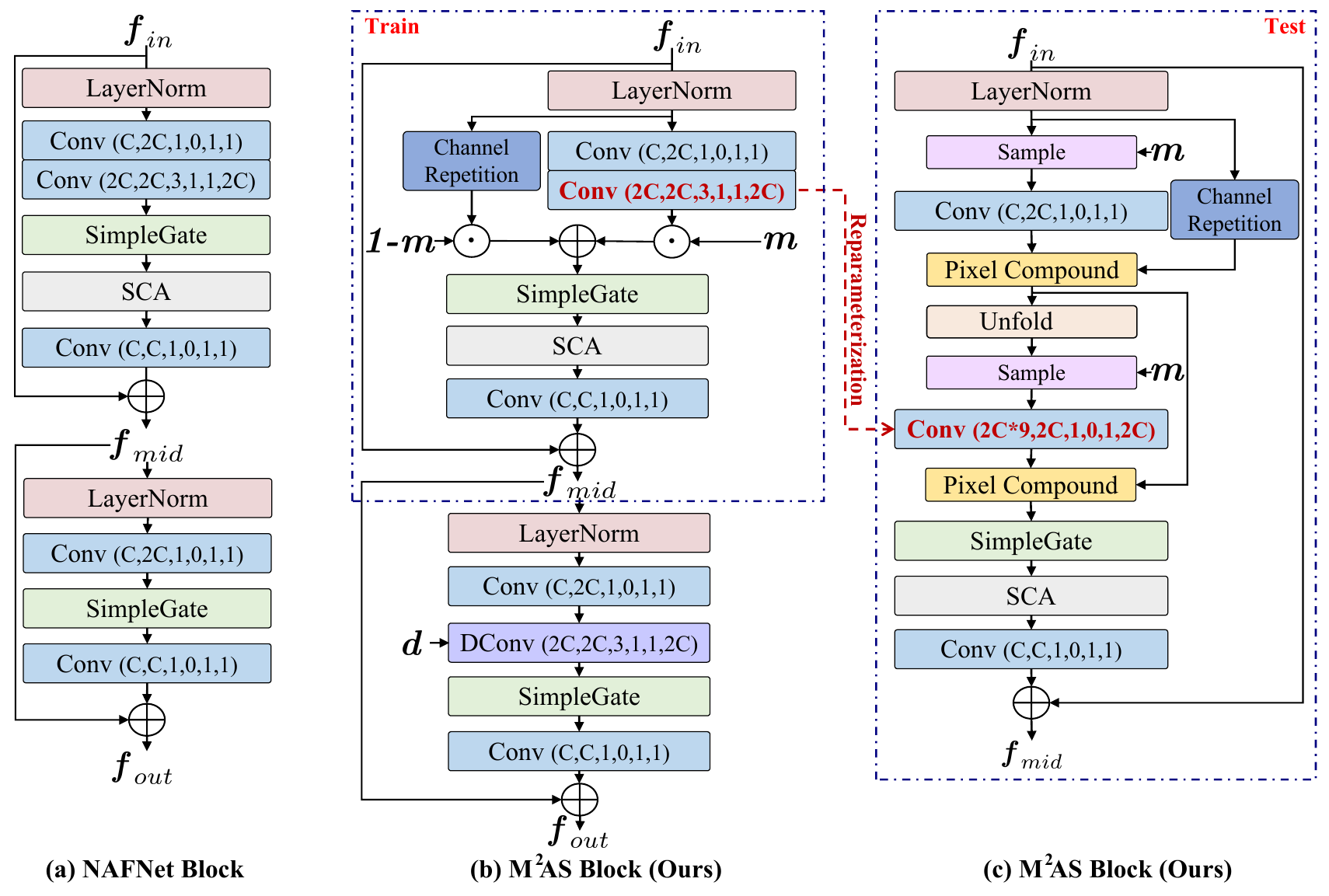}\\
	\end{tabular}
    \vspace{-0.2em}
	\caption{Overview of key differences with the original NAFNet Block. (a) The original NAFNet block~\cite{chen2022simple}. The parameters in $\mathtt{Conv}$ and $\mathtt{DConv}$ denote input channel, output channel, kernel size, padding, stride, and convolution groups, respectively. (b) Our block differs from the original NAFNet block by incorporating mask-aware convolution in the first half of the block. We leverage element-wise multiplication between the features and the mask $\bm m$ to exclude sharp pixels while ensuring backward propagation during training. In the latter half of the block, we use the intra-motion analyzer to calculate the relative displacement $\bm d$ as the intra-motion trajectory and feed it into the motion-aware deformation convolution, indicating the direction and intensity of motion for better deblurring results; details of the motion-aware deformable convolution can be found in Fig.~\ref{fig:dconv}. (c) During testing, we sample features from the blurred regions according to the blur mask $\bm m$ and process them with spatially independent convolution operations (\ie, $1\times 1$ convolution) to reduce computation. To make it applicable to convolutions with kernel sizes larger than 1, we employ reparameterization techniques (reshaping the $3\times 3$ convolution weight from $\bm {W}\in\mathbb{R}^{2C\times 2C\times 3\times 3}$ to $\bm {W}'\in\mathbb{R}^{2C\times (2C\times 9)\times 1\times 1}$) and feature unfolding operations, equivalently replacing the previous $3\times 3$ convolution with $1\times 1$ convolution to meet the needs of pixel-level feature processing. The latter half of the block is the same as the training stage.
	}
	\label{fig:arch}
\end{figure*}
\section{Related Work}
\subsection{Single Image Motion Deblurring Methods}
With the advent of deep learning, data-driven approaches have revolutionized image deblurring. CNN-based methods. Sun~\etal~\cite{sun2015learning} employed a deep network to predict the probabilistic distribution of motion blur from a single image. Pan~\etal~\cite{pan2016blind} found that enforcing the sparsity of the dark channel helps blind deblurring on various scenarios. DeepDeblur~\cite{nah2017deep}, leverage multi-scale convolutional neural networks to restore sharp images from blurred inputs. DeblurGAN~\cite{kupyn2018deblurgan} employs a conditional adversarial framework to generate high-quality deblurred images. Tao~\etal~\cite{tao2018scale} designed a recurrent UNet to deblur progressively. Cho~\etal~\cite{cho2021rethinking} proposed a multi-input and multi-output UNet for fast and accurate image deblurring. Chen~\etal~\cite{chen2022simple} proposed a nonlinear activation free network for image restoration. 
Mao~\etal~\cite{mao2023intriguing} proposed a plug-and-play frequency selection block, which can provide faithful information about the blur pattern, to enhance deblurring performance.
Pham~\etal~\cite{pham2024blur2blur} proposed a blur-to-blur transformation method to convert unseen degradations into known degradations, enabling existing pre-trained models to restore blurred images.
With the advancement of Vision Transformer, numerous Transformer-based methods have been employed for image restoration. Uformer~\cite{wang2022uformer} introduced a locally-enhanced
window Transformer block and a learnable modulator for restoring details. Restormer~\cite{zamir2022restormer} reduced the computational complexity of processing high-resolution images by applying self-attention across channels rather than the spatial dimension. Kong~\etal~\cite{kong2023efficient} developed an efficient frequency domain-based self-attention to replace spatial attention, thereby reducing computational complexity.
Xiao~\etal~\cite{xiao2024unraveling} established a probabilistic representation model that quantifies uncertainties and guides subsequent motion-masked separable attention to deblur.
Mao~\etal~\cite{mao2024adarevd} proposed a reversible structure to separate the degree of degradation from the image content. And based on the spatial variation of the motion blur kernel, a classifier was further introduced to learn the degree of degradation of image patches, enabling them to exit early in different sub-decoders to accelerate the processing process.

\subsection{Region-Aware Image Restoration}
In order to further reduce computational complexity and to process different regions of an image according to their varying degrees of degradation, ClassSR~\cite{kong2021classsr} proposed to categorize different patches of an image based on the difficulty of restoration and employs distinct modules to process patches in each level of difficulty. 
Wang~\etal~\cite{wang2022adaptive} utilized a regressor to predict the incremental capacity of each layer for the patch. This method is capable of reducing computational complexity without introducing additional parameters. 
Jeong~\etal~\cite{jeong2024accelerating} designed a pixel-level classifier and a set of pixel-level upsamplers to distribute computational resources adaptively at the pixel level.
Recently, to address local motion blur in real-world scenarios and to reduce unnecessary computations, Li~\etal~\cite{li2023real} proposed a real local motion blur dataset and bridged the gap between global and local motion deblurring by incorporating gate modules on existing deblur network. 
Then Li~\etal~\cite{li2024adaptive} further proposed adaptive window pruning Transformer blocks to focus deblurring on local regions and reduce computation.
Although Transformer exhibits powerful performance, GPU memory consumption is correspondingly large, and existing methods operate on the patch level, which can still lead to redundancy. In this paper, we adopt a lightweight CNN-based structure and design a pixel-level pruning strategy to further enhance the efficiency of the method. By making decisions at the pixel level, our approach can more precisely allocate computational resources, reducing redundancy and leading to more efficient processing.

\section{Methodology}
To achieve favorable performance with minimal computation, we adopt a simplified structure based on NAFNet~\cite{chen2022simple}. Our method incorporates three critical components at each scale: a mask predictor, an intra-motion analyzer, and multiple mask- and motion-aware simple (M$^2$AS) blocks as shown in Fig.~\ref{fig:arch}(b). We replace NAFNet blocks with our M$^2$AS blocks in NAFNet and add a mask predictor and an intra-motion analyzer at each scale to ensure targeted processing of blurred regions while preserving other areas. Beyond conventional reconstruction loss, we introduce mask loss and reblur loss to regularize the mask predictor and intra-motion analyzer, respectively. 

\subsection{Mask Predictor}
For convenience in expression, we omit the subscript $i$ indicating the scale and uniformly use $\bm f_{in}$ to denote the input features.
For each scale, given the input features $\bm f_{in}$ from the previous scale, the mask predictor employs MLP layers to generate feature embeddings $\bm f'_{in}$ following \cite{rao2021dynamicvit,li2024adaptive}, which are then passed through a softmax activation to produce the predicted probability distribution.
Formally, given the feature map $\bm{f}_{{in}} \in \mathbb{R}^{H \times W \times C}$, we first apply patch embedding to flatten it to $\bm{f}_{{in}} \in \mathbb{R}^{D \times C}$, where $D=H\times W$ denotes the number of pixels. It is then processed through MLP to achieve feature mapping to get $\bm f'_{in}$. 
Subsequently, we derive a global feature by performing channel-wise averaging on $\bm f'_{in}$, which is then utilized to enable per-pixel blur prediction by:
\begin{equation}
\bm{p}=\mathtt{Softmax}(\mathtt{MLP}\left(\mathtt{Concat}\left(\bm f'_{in}, \mathtt{Avg}(\bm f'_{in})\right)\right)), \bm{p}\in \mathbb{R}^{D \times 2}.
\end{equation}
We regard this task as a binary classification dense prediction problem, where each element in ${\bm{p}}_{[*,0]}$ is the probability of blur at that position. $\mathtt{Avg}$ means performing channel-wise averaging on the features.
To facilitate subsequent pixel-wise processing, we need to obtain a binary mask. Following the approach in references \cite{rao2021dynamicvit,li2024adaptive}, we employ Gumbel-Softmax \cite{Jang2017categorical} to generate the mask, ensuring that the gradients can be backpropagated during training:
\begin{equation}
\bm{m}={\mathtt{Gumbel\text{-}Softmax}(\bm{p})}_{[*,0]}, \bm{m}\in \mathbb{R}^{D \times 1}.
\end{equation}
In testing, we compare ${\bm{p}}_{[*,0]}$ with a threshold $\epsilon$:
\begin{equation}
\bm{m}=\left\{\begin{array}{lc}
1, & \text { if }\left({\bm{p}}_{[*,0]}>\epsilon\right) \\
0, & \text { otherwise }.
\end{array}\right.
\end{equation}
The mask $\bm{m}$ will be used to select the blurred regions that require processing, preventing unnecessary impacts on non-blurred areas. $\epsilon$ is set as 0.5 in our experiment.

\subsection{Intra-Motion Analyzer}
The blur mask can only indicate the blurred areas but cannot provide the degree of blur to assist the network in performing more accurate deblurring. Therefore, we propose an intra-motion analyzer to estimate the displacement of each pixel within the exposure time, which will guide the subsequent deblurring block. The structure of the analyzer is very lightweight, consisting of only one convolutional layer. By passing $\bm{f}_{in}$ through it, we can obtain a set of pixel displacements that represent the magnitude of displacements for each pixel during the exposure time. Since blur is caused by signal accumulation during the exposure period, we discretize this process and approximate it as the accumulation of image signals at $N$ time instants. We define the pixel displacement from the midpoint of the exposure time to the start of the exposure time as $\bm{o}^{(t_0)}\in\mathbb{R}^{H\times W\times 2}$, and from the midpoint of the exposure time to the end of the exposure time as $\bm{o}^{(t_{N-1})}\in \mathbb{R}^{H\times W\times 2}$. We use the quadratic function to obtain the other displacements between $\bm{o}^{(t_0)}$ and $\bm{o}^{(t_{N-1})}$, which is confirmed to better approximate real-world motion \cite{xu2019quadratic}. It can be written as:
\begin{equation}
\begin{aligned}
\bm{o}^{(t_n)} & =\frac{\bm{o}^{(t_0)}+\bm{o}^{(t_{N-1})}}{2}\left(\frac{2 n}{N-1}-1\right)^2 \\
& +\frac{\bm{o}^{(t_{N-1})}-\bm{o}^{(t_0)}}{2}\left(\frac{2 n}{N-1}-1\right), n=0, \ldots, N-1
\end{aligned}
\end{equation}
It is not difficult to calculate that when $n=\frac{N-1}{2}$, $\bm{o}^{(t_\frac{N-1}{2})}$ is $\bm 0$. $\{\bm{o}^{(t_n)}\}^{N-1}_{n=0}$ not only contains the magnitude of motion offset but also includes the motion direction, which can effectively assist in deblurring.

\begin{figure} 
	\centering
	\begin{tabular}{c}
		\includegraphics[width=\linewidth]{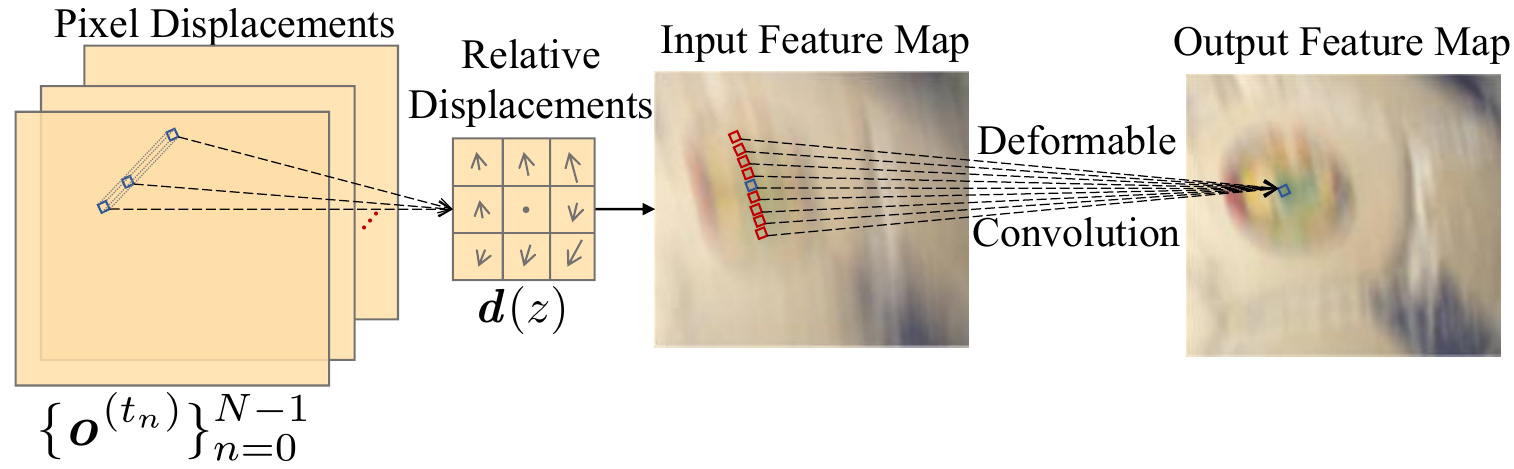}\\
	\end{tabular}
	\caption{The pipeline of our motion-aware deformable convolution involves calculating the relative displacement $\bm{d}(z)$ for any spatial pixel position $z$ based on the pixel displacements generated by the intra-motion analyzer. This displacement $\bm{d}(z)$ replaces the traditional offset in deformable convolution, allowing us to directly incorporate pixels related to the motion trajectory of pixel $z$ during exposure time to compute the corresponding sharp value. 
	}
	\label{fig:dconv}
\end{figure}
\subsection{Mask- and Motion-Aware Simple Block}
Unlike global motion blur which affects the entire image, local motion blur is confined to specific regions within the image. This type of local blur is typically the result of object movement captured by a stationary camera. Applying global deblurring methods to images with local motion blur inevitably introduces undesirable artifacts in originally clear areas~\cite{li2024adaptive}. Therefore, we propose a motion-aware adaptive pixel pruning strategy based on the estimated blur mask $\bm m$ to distinctly differentiate and process blurred objects from the background. 
Our mask- and motion-aware simple block ensures efficiency and targets only the blurred regions for processing, further reducing computation. Additionally, we introduce blur motion information $\{\bm{o}^{(t_n)}\}^{N-1}_{n=0}$ to indicate the degree of blur and the direction of movement, thereby assisting the network in the deblurring process. The architecture is shown in Fig.~\ref{fig:arch}(b). The core modifications primarily include mask-aware convolution and motion-aware deformable convolution.

\begin{table*} 
	\centering
    \caption{Quantitative comparisons on the local and global deblurring datasets. “$\text{PSNR}_{w}$”, “$\text{SSIM}_{w}$” denote weighted PSNR, weighted SSIM. \textbf{Bold} and \underline{underline} are used to indicate top $1^\text{st}$ and $2^\text{nd}$, respectively.}
    \label{table:local_global_blur}
	\begin{tabular}{c|cccc|ccc}
		\hline
        
        \hline	
		\multirow{2}{*}{Methods} & \multicolumn{4}{|c}{ReLoBlur (Local Blur)}& \multicolumn{2}{|c}{GoPro (Global Blur)} \\ 
		\cline{2-7}
		& PSNR & SSIM & $\text{PSNR}_{w}$& $\text{SSIM}_{w}$ & PSNR & SSIM\\ 

		\hline
        DeepDeblur~\cite{nah2017deep} &  33.021   &  0.8904  &  28.289  & 0.8398  &  28.552  & 0.8623 \\
        HINet~\cite{chen2021hinet}  &  34.400   &  0.9160  &  29.291  & 0.8489   &  28.658 & 0.8674  \\
        MIMOUNet+~\cite{cho2021rethinking} & 34.421  &  0.9239 &  28.801   & 0.8746  &  30.228  & 0.9026\\
        NAFNet~\cite{chen2022simple}  & 34.584   &  0.9244 &  28.633 & 0.8751  &  32.402  &  0.9338 \\
		Uformer~\cite{wang2022uformer}  &  34.831   &  0.9274  &  29.265  & 0.8806 &  32.316  & 0.9305 \\
		Restormer~\cite{zamir2022restormer} & 34.670  &  0.9228 &  29.436 &  0.8793  & 32.452 & \underline{0.9365}\\
		LBAG~\cite{li2023real} &  34.655   &  0.9257  &  29.320   &  0.8810  &  30.702 & 0.9098 \\
		LMD-ViT~\cite{li2024adaptive} &  \underline{35.394} & \underline{0.9280} &  \underline{30.240}  & \underline{0.8929}  & \underline{32.500} & 0.9312 \\
		Ours &  \textbf{35.647}  & \textbf{0.9302}  &   \textbf{31.036}  & \textbf{0.8997}  & \textbf{33.055}  & \textbf{0.9406}\\

		\hline
		
		\hline
	\end{tabular}
	
\end{table*}

\begin{table*} 
	\centering
    \caption{Efficiency comparisons on the ReLoBlur dataset. “Time” and “GPU memory” denote the inference time and the GPU memory consumption, respectively. }
    \label{table:efficiency}
	\begin{tabular}{c|cccc}
		\hline
        
        \hline	
		Methods & Params (M) & FLOPs (T) & GPU memory (GB) & Times (s) \\

		\hline
        DeepDeblur~\cite{nah2017deep} &  11.72 & 15.851 &  \textbf{4.6} &  7.234 \\
        HINet~\cite{chen2021hinet}  &  88.67   &  8.162  & 14.6 & 0.799  \\
        MIMOUNet+~\cite{cho2021rethinking} & \textbf{16.11} &  7.850 &   9.2  & 1.090\\
        NAFNet~\cite{chen2022simple}  &  \underline{17.11}  & \underline{0.764}  &  \underline{5.0} & \textbf{0.569} \\
		Uformer~\cite{wang2022uformer}  &  50.88   &  4.375  &  35.3  & 3.592  \\
		Restormer~\cite{zamir2022restormer} &  26.13 &  6.741 & 37.2  & 3.536 \\
		LBAG~\cite{li2023real} & \textbf{16.11} &  7.852  &  9.2   & 1.120 \\
		LMD-ViT~\cite{li2024adaptive} & 54.55 & 1.485 &  28.0  &  1.284 \\
		Ours  &  17.44  &  \textbf{0.759}  &  5.8  &  \underline{0.790}  \\

		\hline
		
		\hline
	\end{tabular}
\end{table*}
\subsubsection{Mask-Aware Convolution}
To ensure that the network focuses on processing blurred regions, we perform element-wise multiplication between the input features and the mask $\bm m$ during training to exclude pixels that do not require processing. Finally, these pixels are combined with the original input features to form the final output features $\bm{f}_{mid}$:
\begin{equation}
\begin{aligned}
    \bm{f}_{1} &= \mathtt{LN}(\bm{f}_{in}), \\
	\bm{f}_2 &= \mathtt{Conv}(\bm{f}_{1})\odot \bm m + \mathtt{Rep}(\bm{f}_{1})\odot (\bm 1 - \bm m), \\
    \bm{f}_3 &= \mathtt{SCA}(\mathtt{SG}(\bm{f}_2)), \\
    \bm{f}_{mid} &= \mathtt{Conv}_{1\times 1}(\bm{f}_3) + \bm{f}_{in},
\end{aligned}
\end{equation}  
where $\mathtt{LN}$ is layer normalization, $\mathtt{SCA}$ and $\mathtt{SG}$ denote simplified channel attention and simple gate used in \cite{chen2022simple}. $\mathtt{Rep}$ is channel repetition operation. $\mathtt{Conv}$ contains one $1\times 1$ convolution layer and one $3\times 3$ depth-wise convolution layer. $\mathtt{Conv}_{1\times 1}$ is only one $1\times 1$ convolution layer.

However, merely using element-wise multiplication only allows the network to process blurred areas but does not reduce the computation. Therefore, during the testing phase, we employ a pixel pruning strategy to sample pixels that require processing based on the mask $\bm m$ to obtain a new feature with size $\mathbb{R}^{Q\times 1\times C}$, allowing the network to focus solely on blurred pixels, and thus reducing computational costs. $Q$ refers to the number of pixels with a value of 1 in the mask. Since $1\times 1$ convolution is spatially invariant, it can be applied directly to the sampled features. For $3\times 3$ convolutions, we reparameterize them with $1\times 1$ convolutions. Specifically, for the original weights $\bm {W}\in\mathbb{R}^{C_{out}\times C_{in}\times 3\times 3}$, we reshape them into $\bm {W}'\in\mathbb{R}^{C_{out}\times (C_{in}\times 9)\times 1\times 1}$.
Consequently, we apply an unfold operation with the kernel size of 3 to the features $\bm{f}_{1} \in \mathbb{R}^{H\times W \times C_{in}}$, which expands them into features $\bm{f}'_{1} \in \mathbb{R}^{H\times W \times (C_{in}\times 3 \times 3)}$ that maintains the same receptive field as the $3\times 3$ convolution operation while ensuring that the input channel of the features matches the input channel of the reparameterized convolutional layers.
Lastly, the processed regions are placed back into the original feature map according to the mask to obtain the final output features $\bm{f}_{2}$. It is not difficult to calculate that the computation of this operation only accounts for a small fraction of the original, $\frac{Q}{H\times W}$. When processing high-resolution images and the blurred areas are relatively small, the advantage becomes particularly evident. Our mask convolution is theoretically applicable to any convolution and we analyze the impact of replacing different convolution layers with mask convolution in Sec~\ref{sec:abl}.

\subsubsection{Motion-Aware Deformable Convolution}
Since motion blur is caused by the accumulation of signals over the exposure time, the convolution filter should have a similar shape to the trajectory to ensure that the pixels involved in the computation are related to the current position as shown in Fig.~\ref{fig:dconv}. Therefore, we employ motion-aware deformable convolution. Unlike standard deformable convolutions, we calculate the offsets in the deformable convolution based on the previously obtained pixel displacements. Taking the kernel size of $3\times 3$ as an example, a total of 9 weight offsets are required. We need to calculate the deformable offsets based on the midpoint of the exposure time $t_\frac{N-1}{2}$, which requires $N=N_1=9$. For a standard convolution kernel, its relative positions can be represented as (-1,-1),(-1,0),\ldots,(0,1),(1,1). Adding these to $\{\bm{o}^{(t_n)}\}_{n=0}^{N_1-1}$, we obtain the relative displacements $\bm d\in\mathbb{R}^{H\times W \times (2\times 9)}$ for the deformable convolution. We add this deformable convolution between the $1\times 1$ convolution layer and the simple gate as shown in Fig.~\ref{fig:arch}(b). The entire process can be represented as follows:
\begin{equation}
\begin{aligned}
    \bm{f}_{4} &= \mathtt{Conv}_{1\times 1}(\mathtt{LN}(\bm{f}_{mid})), \\
	\bm{f}_5 &= \mathtt{SG}(\mathtt{DConv}(\bm{f}_{4}, \bm d)), \\
    \bm{f}_{out} &= \mathtt{Conv}_{1\times 1}(\bm{f}_5) + \bm{f}_{mid},
\end{aligned}
\end{equation}
where $\mathtt{DConv}$ is the motion-aware deformable convolution. To avoid introducing too many additional parameters, we implement a depth-wise deformable convolution.

\begin{figure*} 
	\centering
	\begin{tabular}{c}
		\includegraphics[width=0.85\linewidth]{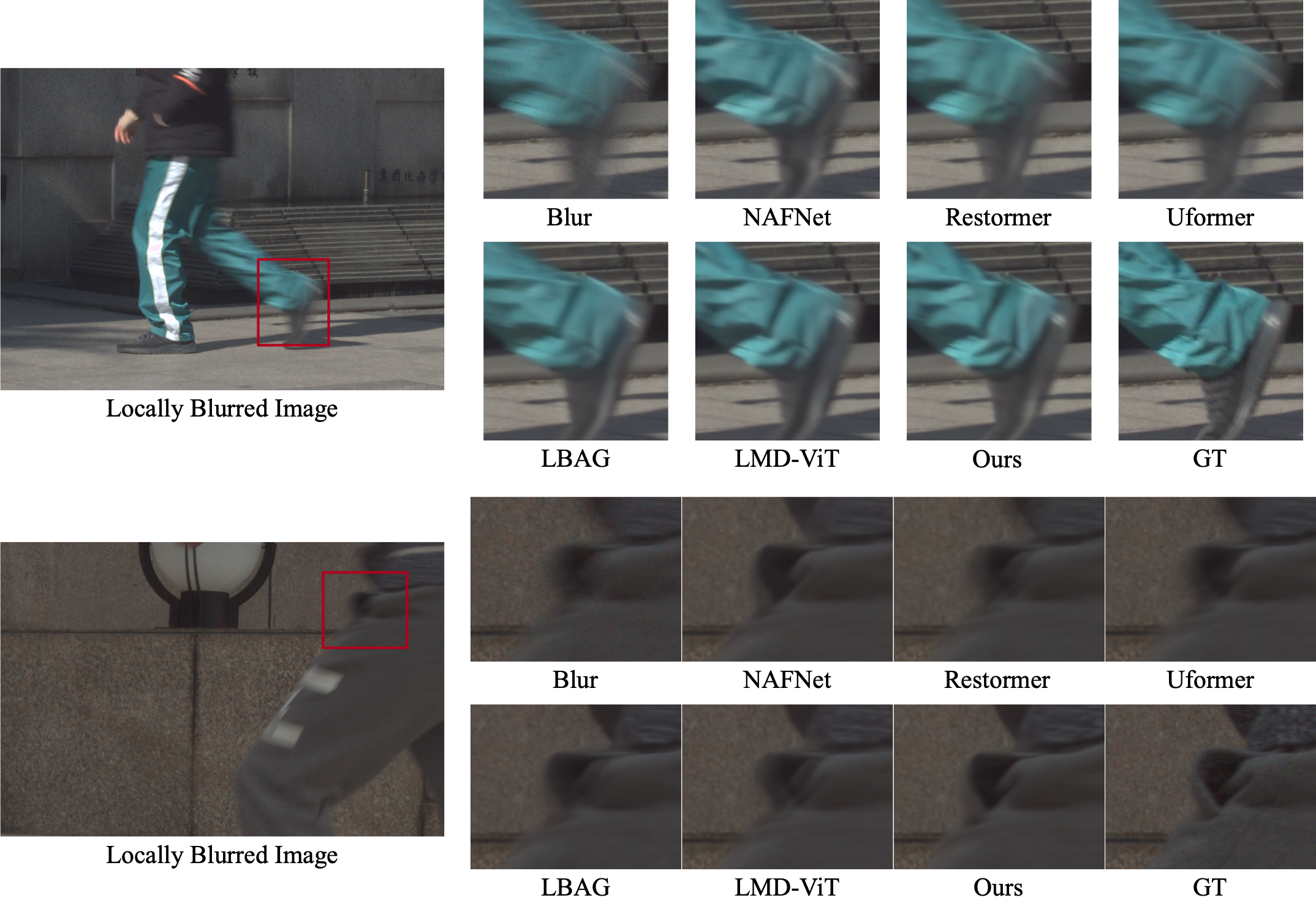}\\
	\end{tabular}
	\caption{Visual comparison with the state-of-the-art deblurring methods on ReLoBlur~\cite{li2023real}.
	}
	\label{fig:local}
\end{figure*}

\subsection{Loss Functions}
For joint learning, we adopted a loss function composed of three parts: an offset prediction loss, a blur mask prediction loss, and a reconstruction loss to form the final loss:
\begin{equation}
	\mathcal{L}=\mathcal{L}_{recon} + \lambda \mathcal{L}_{mask}+\mathcal{L}_{offset}.
\end{equation} 
\subsubsection{Reconstruction Loss}
\begin{equation}
	\mathcal{L}_{recon}=\mathtt{L}_{1}({\bm{y}}, \bm{y}^{gt}) + \mathtt{SSIM}({\bm{y}}, \bm{y}^{gt}) +  \gamma \mathtt{FFT}({\bm{y}}, \bm{y}^{gt}),
\end{equation}
where $\bm{y}$ is the output of our network and $\bm{y}^{gt}$ denotes the sharp ground-truth.
Following the setting in~\cite{li2023real, li2024adaptive}, we also utilize $\mathtt{L}_{1}$ loss, SSIM loss $\mathtt{SSIM}$, and frequency reconstruction loss $\mathtt{FFT}$ as our reconstruction loss. And we set $\lambda=0.01$ and $\gamma=0.1$ according to~\cite{li2023real, li2024adaptive}.

\subsubsection{Mask Prediction Loss}
\begin{equation}
	\mathcal{L}_{mask}=\sum^{S}_{i=1}\mathtt{Cross\text{-}Entropy}({{\bm{m}_i}}, \bm{m}^{gt}_i),  
\end{equation} 
where $S$ is the number of scales in the network, and $\bm{m}^{gt}_i$ represents the ground-truth mask in the $i$-th scale.
\subsubsection{Offset Prediction Loss}
\begin{equation}
    \mathcal{L}_{offset}=\sum^{S}_{i=1}(\mathtt{Reblur}({\{\bm{o}^{(t_n)}_i\}_{n=0}^{N-1}}, \bm{x}_i, \bm{y}^{gt}_i) + \alpha\sum^{N-1}_{n=0}\mathtt{TV}(\bm{o}^{(t_n)}_i)),  
\end{equation}
where $\bm{x}_i$ is the blurred image downsampled to match the resolution of the $i$-th scale. $\mathtt{TV}$ is the total variation loss to ensure smooth pixel displacement maps.
\begin{equation}
\mathtt{Reblur}({\{\bm{o}^{(t_n)}_i\}_{n=0}^{N-1}}, \bm{x}_i, \bm{y}^{gt}_i)= \mathtt{MSE}(\sum^{N-1}_{n=0}\mathtt{Warp}(\bm{y}^{gt}_i, \bm{o}^{(t_n)}_i), \bm{x}_i)
\end{equation}  
where $N=N_2$ is only used to constrain $\bm{o}^{(t_0)}$ and $\bm{o}^{(t_{N-1})}$ in training.
Since $\bm{y}^{gt}_i$ typically represents the sharp frame at the midpoint of the exposure time, we can generate potential sharp frames for other instants based on the pixel displacements from the midpoint to other times $\bm{o}^{(t_n)}_i$ within the exposure period. By utilizing a forward warping operation in \cite{niklaus2020softmax}, we can synthesize the blurred frame through signal accumulation of these warped sharp frames and constrain it with the input blurred frames to achieve supervision of the pixel displacements.
\begin{figure} 
	\centering
	\begin{tabular}{c}
		\includegraphics[width=\linewidth]{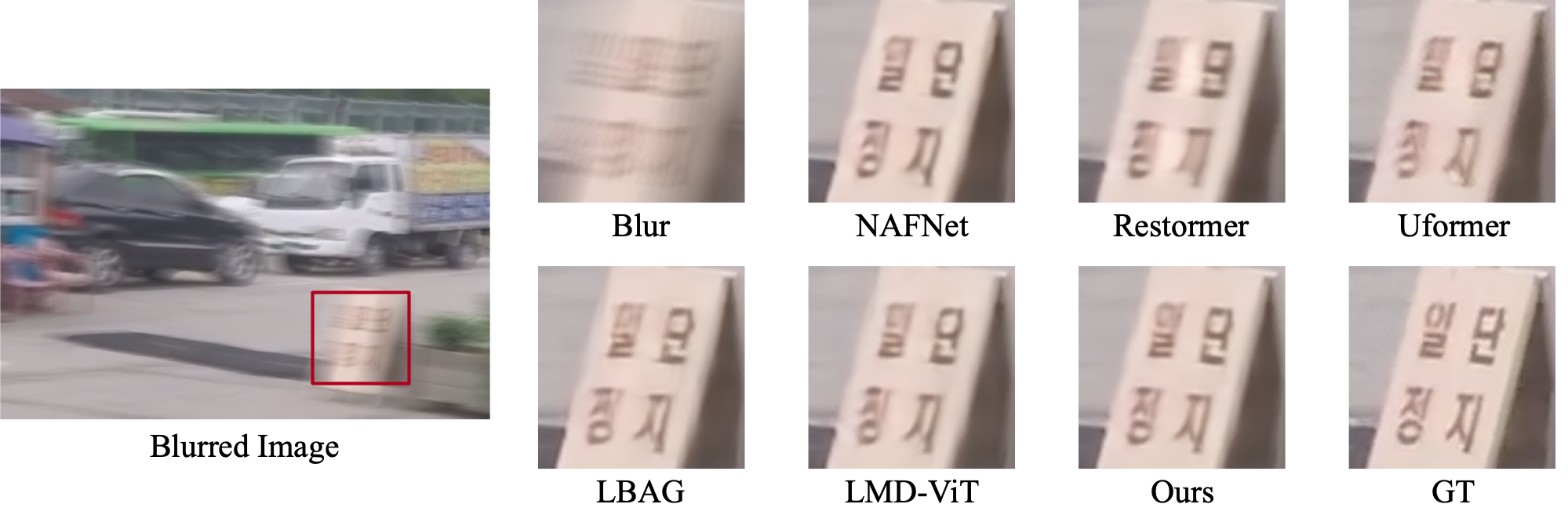}\\
	\end{tabular}
	\caption{Visual comparison with the state-of-the-art deblurring methods on GoPro~\cite{nah2017deep}. 
	}
	\label{fig:global}
\end{figure}
\section{Experiments}
\subsection{Datasets}
Following~\cite{li2024adaptive}, our method is trained on the GoPro dataset~\cite{nah2017deep} and the ReLoBlur dataset~\cite{li2023real} with a 1:1 sampling rate to address both local and global motion deblurring issues. 
The ReLoBlur dataset~\cite{li2023real} is a real-world dataset captured using a beam-splitting photography system, which yields paired blur-sharp images. This dataset encompasses a training set of 2010 images and a test set of 395 images, both with the full image size of 2152 $\times$ 1436. \cite{li2024adaptive} employed an enhanced annotation scheme to generate more accurate local blur masks instead~\cite{li2023real}. In contrast, the GoPro dataset contains mainly global blur, so we utilize $\mathbf{1}$ as the blur mask. 
We use the Peak Signal-to-Noise Ratio (PSNR) and Structural Similarity Index (SSIM)~\cite{wang2004image} metrics to evaluate the deblurred results on the global blur dataset. For local blur, following the approach in~\cite{li2023real,li2024adaptive}, we additionally calculate the weighted PSNR and weighted SSIM based on the blur mask to represent the quality of restoration in the blurred regions. To present model efficiency, we also report the comparison on the inference time, Floating Point Operations Per Second (FLOPs), GPU memory consumption and model parameters.

\subsection{Implementation Details}
We train M$^2$AENet using AdamW optimizer~\cite{kingma2014adam} with the momentum terms of (0.9, 0.9), the weight decay of $1\times 10^{-3}$, a batch size of 8, and an initial learning rate of $1\times 10^{-3}$ that is gradually lowered to $1\times 10^{-6}$ by cosine annealing~\cite{loshchilov2016sgdr}.
We employ a blur-aware patch cropping strategy~\cite{li2023real} and set the patch size to $256$.
All training and testing experiments are conducted using PyTorch on an NVIDIA RTX A6000 GPU. To ensure the fairness of the experiments, we retrain all state-of-the-art methods on the combined dataset with the same cropping strategy. The model configurations of the compared deblurring methods follow their original settings.

\begin{figure} 
	\centering
	\begin{tabular}{c}
		\includegraphics[width=\linewidth]{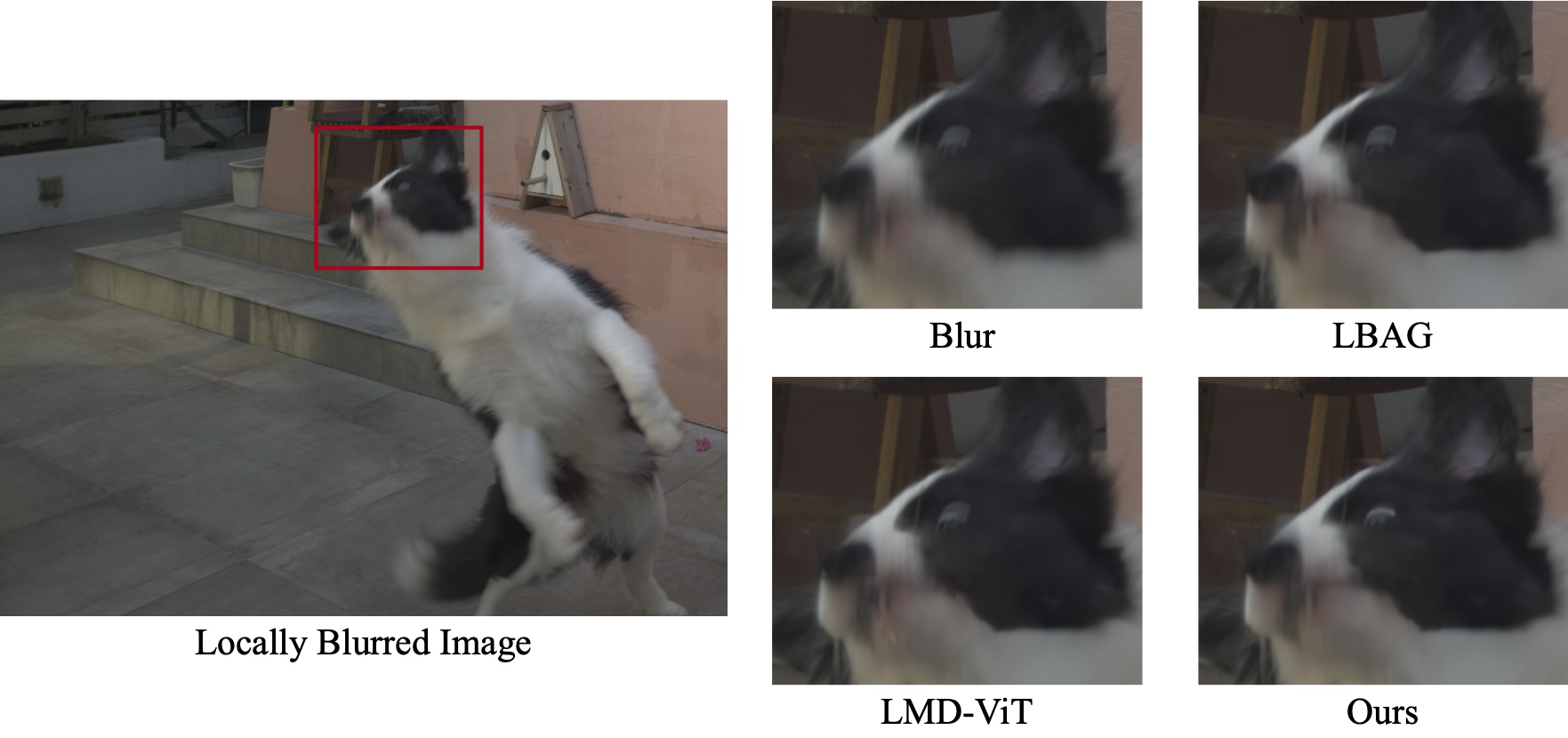}\\
	\end{tabular}
	\caption{Visual comparison with the state-of-the-art deblurring methods on real-world locally blurred images. 
	}
	\label{fig:real}
\end{figure}
\subsection{Comparison with State-of-the-art Methods}
We compare M$^2$AENet with state-of-the-art deblurring methods on the ReLoBlur dataset and the GoPro dataset. We choose methods from two categories: 1) global deblurring methods, including DeepDeblur~\cite{nah2017deep}, HINet~\cite{chen2021hinet}, MIMOUNet+~\cite{cho2021rethinking}, NAFNet~\cite{chen2022simple}, Uformer~\cite{wang2022uformer} and Restormer~\cite{zamir2022restormer} and 2) local deblurring methods, including LBAG~\cite{li2023real} and LMD-ViT~\cite{li2024adaptive}. 

As shown in Table~\ref{table:local_global_blur}, our method achieves optimal performance on both local blur and global blur datasets. On the local blur dataset ReLoBlur, our method outperforms the latest local deblurring methods by 0.25dB and achieves a significant advantage of nearly 0.8dB in the $\text{PSNR}_{w}$ metric. This demonstrates the effectiveness of our method in removing local motion blur while preserving clear regions. For global blur, our method shows a 0.55dB improvement, indicating its ability to handle both local and global blur effectively. This is attributed to our mask- and motion-aware simple block, which processes blurred regions and extracts motion magnitude and direction to guide image restoration.
Visual comparisons shown in Fig.~\ref{fig:local}, our method shows clear advantages, such as better restoration of details like wrinkles on pants and pockets. Other methods, especially those for global blur, fail to effectively restore locally blurred objects due to the lack of blur detection and localization. Fig.~\ref{fig:global} shows our method effectively removes global blur, restoring text on a sign to a legible state, achieving a visual effect closer to the ground-truth (GT).
To validate our model on real-world data, we captured locally blurred RGB images with $4096\times 3072$ resolution and compared our method with existing local deblurring methods in Fig.~\ref{fig:real}. For dogs exhibiting motion blur caused by jumping during photography, our approach more effectively eliminates blur artifacts while restoring critical details such as eyes and ears, further proving its effectiveness on real-world data.

\begin{table} 
	\centering
    \caption{Ablation analysis of each component in our M$^2$AENet. See the text for the details of different variants.}
    \label{table:abl}
	\begin{tabular}{c|ccccc}
		\hline
        
        \hline	
		Variants & PSNR & SSIM & $\text{PSNR}_{w}$& $\text{SSIM}_{w}$ & FLOPs (T)\\ 

		\hline
        1)  &  34.584   &  0.9244  &  28.633  & 0.8751  &  0.764 \\
        2)  &  35.091   &  0.9271  &  29.591  & 0.8889 & 0.995   \\
        3) & 35.021  &  0.9269 &  30.081   & 0.8916 &  0.528 \\
        4)  & 35.647   &  0.9302 &  31.036 & 0.8997  &  0.759 \\

		\hline
		
		\hline
	\end{tabular}
\end{table}

Model size and computational efficiency are critical for real-world applications. We comprehensively compare our method with SOTAs in terms of parameters, FLOPs, inference time, and GPU memory usage, demonstrating significant advantages. Built on NAFNet with the added mask predictor and intra-motion analyzer, our method has 0.33M more parameters than NAFNet but achieves lower computational costs and similar GPU memory usage due to adaptive pixel pruning, as shown in Table~\ref{table:efficiency}. Compared to the top-performing LMD-ViT, which uses a Transformer-based architecture and consumes more memory for high-resolution images, our method reduces FLOPs by 49\%, inference time by 38.5\%, and GPU memory usage to just 21\% of LMD-ViT. This highlights the balance of our method between performance and efficiency.

\subsection{Ablation Studies}\label{sec:abl}
\begin{figure} 
	\centering
	\begin{tabular}{c}
		\includegraphics[width=0.55\linewidth]{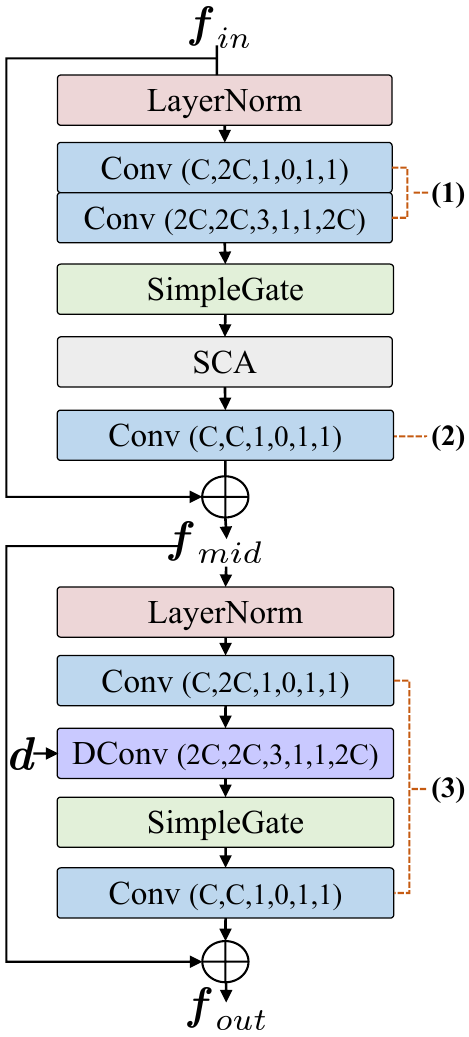}\\
	\end{tabular}
	\caption{The position where mask-aware convolution can be applied. 
	}
	\label{fig:maconv_pos}
\end{figure}
In this section, we first analyze the effectiveness of the mask-aware convolution and motion-aware deformable convolution in Table~\ref{table:abl}, and specifically examine the impact of the positions where mask-aware convolution is applied as shown in Fig.~\ref{fig:maconv_pos} and Table~\ref{table:maconv_pos}.

We design the following variations: 1) Removing mask-aware convolution and motion-aware deformable convolution, along with the corresponding mask predictor and intra-motion analyzer, reduces the design to the original NAFNet block.
2) Removing only mask-aware convolution and the mask predictor, while replacing them with standard convolution.
3) Removing motion-aware deformable convolution and the corresponding intra-motion analyzer.
4) Retaining all components, which constitutes our complete method.
As shown in Table~\ref{table:abl}, a comparative analysis between variants 4) and 2) reveals that the removal of mask-aware convolution severely compromises local deblurring performance (as evidenced by reduced $\text{PSNR}_{w}$ metrics). This stems from the diminished capacity of the model for spatial differentiation, resulting in a reversion to conventional global deblurring strategies that lack precise regional processing capabilities.
The performance comparison between variants 4) and 3) demonstrates that eliminating motion-aware deformable convolution induces significant deterioration in global restoration quality (as evidenced by reduced PSNR metrics). While retaining basic local deblurring functionality, the absence of motion-adaptive operations prevents dynamic adjustment to the pixel displacements, thereby impairing directional blur rectification.
From this analysis, it is clear that both components are essential. Only by integrating mask-aware and motion-aware convolutions can optimal deblurring performance be achieved.

We analyze the application positions of mask-aware convolution within the block in Table~\ref{table:maconv_pos}. While our proposed adaptive pixel pruning strategy can theoretically be applied to any convolution layer, it offers greater benefits for layers with a kernel size larger than 1, since the computational overhead of larger kernels grows substantially for the same number of channels.
In position (1) of Fig.~\ref{fig:maconv_pos}, replacing the convolution with mask-aware convolution ensures targeted processing of blurred regions while reducing computation. For positions (2) and (3), as shown in Table~\ref{table:maconv_pos}, replacing additional convolution layers within the block with mask-aware convolution does not yield further benefits. Although it reduces computational costs, it also degrades overall model performance. Therefore, we only replace the convolution layer at position (1) with mask-aware convolution to balance computational efficiency and model effectiveness.
\begin{table} 
	\centering
    \caption{Ablation analysis of applying mask-aware convolution at different positions within the block. The detailed positions are illustrated in Fig.\ref{fig:maconv_pos}.}
    \label{table:maconv_pos}
	\begin{tabular}{ccc|ccccc}
		\hline
        
        \hline	
		(1) & (2) & (3) & PSNR & SSIM & $\text{PSNR}_{w}$& $\text{SSIM}_{w}$  & FLOPs (T) \\ 

		\hline
        \Checkmark  & \Checkmark  & \Checkmark  &  35.163   &  0.9264  &  30.541  & 0.8941  & 0.568\\
        \Checkmark  & \Checkmark  & \ding{55}  &  35.519   &  0.9293  &  30.801  & 0.8973  &  0.694 \\
        \Checkmark  & \ding{55}  & \Checkmark & 35.372  &  0.9289 &  30.577   & 0.8951 &  0.633\\
        \Checkmark & \ding{55} & \ding{55}  & 35.647   &  0.9302 &  31.036 & 0.8997   &  0.759\\

		\hline
		
		\hline
	\end{tabular}
\end{table}

\section{Conclusion}
This paper proposes an end-to-end joint optimization method, M$^2$AENet, which enables fine-grained pixel-level processing of blurred regions and analyzes the motion trajectories of blurred objects. To achieve this, we use masks predicted by a mask predictor to exclude sharp regions and adopt an adaptive pixel pruning strategy during testing to further reduce computational costs. Additionally, we leverage an intra-motion analyzer to determine the relative displacements of each pixel during the exposure time, feeding the displacements into deformable convolution to explicitly guide deblurring based on motion trajectories. Our experiments demonstrate the advantages of our method in terms of performance, efficiency, and generalization.

\section*{Acknowledgments}  

This work was supported in part by the National Natural Science Foundation of China (62172127, U22B2035, 62222608 and 62436002), the Natural Science Foundation of Heilongjiang Province (YQ2022F004), and in part by Tianjin Natural Science Funds for Distinguished Young Scholar (23JCJQJC00270).

\clearpage
\bibliographystyle{ACM-Reference-Format}
\bibliography{sample-base}

\clearpage
\appendix
\section{M$^2$AENet architecture and model hyper-parameters}
Our proposed M$^2$AENet is a U-shaped network built upon NAFNet with one
in-projection layer, four encoder stages, one bottleneck stage, four decoder stages, and one out-projection layer. Skip connections are set up between each encoder and decoder stage.
In the encoder, as the stage increases, the spatial resolution of the feature halves, and the channel number doubles. In the decoder, this process is symmetric, with the spatial resolution of the feature doubling and channel number halving. We set the number of blocks per stage consistent with NAFNet~\cite{chen2022simple}: [1, 1, 1, 28], 1, [1, 1, 1, 1], with a base channel number of 32. The last stage of the encoder has more blocks due to its smallest spatial resolution and lower computational load, which is key for efficiency. 
We conducted systematic architectural exploration by implementing encoder, bottleneck, and decoder blocks with configurations [2, 4, 4, 8], 4, and [8, 4, 4, 2], respectively. However, empirical evaluations revealed this architecture failed to yield performance improvements while substantially increasing resource demands: GPU memory consumption escalated to 21 GB (3.62$\times$ expansion) and computational complexity rose to 2.1 TFLOPs (2.77$\times$ magnification). Consequently, we maintained the block configuration of NAFNet to preserve computational efficiency. To ease training, we added the mask predictor and intra-motion analyzer only to the fourth stage of the encoder.
\begin{table} 
	\centering
    \caption{Ablation study of the pixel displacement generation on ReLoBlur.}
    \label{table:interp}
	\begin{tabular}{c|cccc}
		\hline
        
        \hline	
		 & PSNR & SSIM & $\text{PSNR}_{w}$& $\text{SSIM}_{w}$ \\ 

		\hline
        
        Linear  &  35.493   &  0.9271  &  30.891  & 0.8889   \\
        \textbf{Quadratic} &  35.647   &  0.9302 &  31.036 & 0.8997 \\
        Learnable  &  35.651   &  0.9303  &  31.045  & 0.8999  \\

		\hline
		
		\hline
	\end{tabular}
\end{table}
\section{Pixel Displacement Generation}

In the main manuscript, we introduced the intra-motion analyzer, which generates pixel displacements from the midpoint of exposure time to the start as $\bm{o}^{(t_0)}\in\mathbb{R}^{H\times W\times 2}$, and to the end as $\bm{o}^{(t_{N-1})}\in \mathbb{R}^{H\times W\times 2}$. We then interpolate the other pixel displacements with a quadratic function. In this section, we explore alternative methods such as the linear function and learning-based approaches. The linear function to obtain the other displacements between $\bm{o}^{(t_0)}$ and $\bm{o}^{(t_{N-1})}$ can be written as:
\begin{equation}
\bm{o}^{(t_n)}=\left\{\begin{array}{lc}
(1-\frac{2 n}{N-1})\bm{o}^{(t_0)}, & 0\leq n\textless \frac{N-1}{2} \\
(\frac{2 n}{N-1}-1)\bm{o}^{(t_{N-1})}, & \frac{N-1}{2}\textless n\leq {N-1}
\end{array}\right.
\end{equation}
For the learning-based approach, we directly generate all pixel displacements by feeding $\bm{f}_{in}$ through the intra-motion analyzer to generate $\bm{o}\in\mathbb{R}^{H\times W\times {2 N}}$ and then we divide it into $N=N_1=9$ time steps and obtain $\{\bm{o}^{(t_n)}\}^{N_1-1}_{n=0}$.
As shown in Table~\ref{table:interp}, the learning-based method produces better results but requires generating more pixel displacements, which introduces additional parameters. Since it does not offer significant performance improvements over quadratic interpolation, we generate only two pixel displacements and use quadratic interpolation for the rest displacements in our experiments.

\begin{table} 
	\centering
	\caption{Ablation study of the time instants $N_2$ on ReLoBlur.}
	\label{table:off_num}
	\begin{tabular}{c|cccc}
		\hline
		
		\hline	
		& PSNR & SSIM & $\text{PSNR}_{w}$& $\text{SSIM}_{w}$ \\ 
		
		\hline
		$N_2$=5  &  35.111   &  0.9269  &  30.291  & 0.8930   \\
		\textbf{$N_2$=9} &  35.647   &  0.9302 &  31.036 & 0.8997 \\
		$N_2$=15  &  35.651   &  0.9302  &  31.040  & 0.8998  \\
		$N_2$=30  &  35.652   &  0.9302  &  31.040  & 0.8998  \\
		
		\hline
		
		\hline
	\end{tabular}
\end{table}

Next, we discuss the configuration of the time instants $N_2$ in $\mathcal{L}_{offset}$. Increasing $N_2$ will generate more sharp frames through warping operations. As demonstrated in~\cite{nah2019ntire}, synthesizing blur using higher frame-rate sharp frames reduces artifacts and better approximates real-world blur patterns. Table~\ref{table:off_num} reveals that no significant performance gains are observed when $N_2\textgreater 15$. Notably, $N_2=9$ achieves comparable performance to $N_2=15$ while requiring lower computational costs. Considering that deformable convolutions necessitate $N_2\geq 9$ for the offset calculations of deformable convolution, we set $N_2=9$ to balance computational efficiency and restoration quality.

\section{Limitations and Discussions}
\begin{figure} 
	\centering
	\begin{tabular}{c}
		\includegraphics[width=\linewidth]{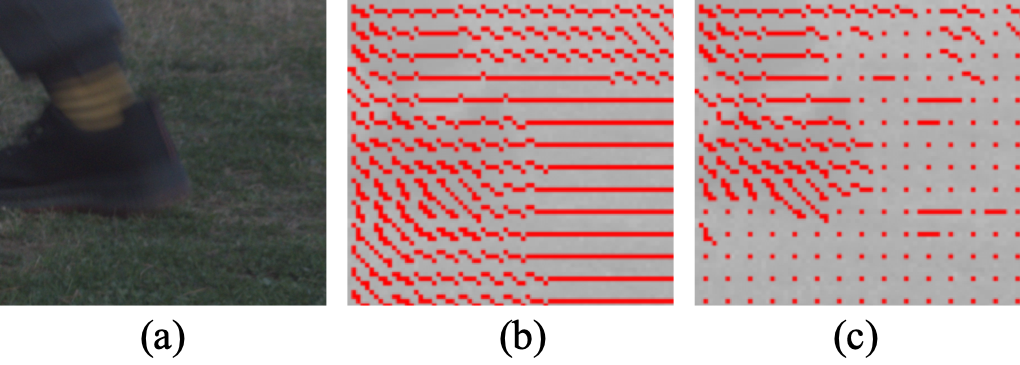}\\
	\end{tabular}
	\caption{(a) Locally Blurred Image. (b) Motion trajectory w/o $\mathcal{L}_{mask}$. (c) Our motion trajectory.  
	}
	\label{fig:lim}
\end{figure}
Our method relies on both a mask predictor and an intra-motion analyzer to jointly localize blurred regions and estimate motion trajectories. 
While the concurrent adoption of these two components may appear redundant, our ablation studies reveal that relying solely on motion trajectory estimation (e.g., through $\mathcal{L}_{offset}$ supervision) proves inadequate for accurate blur identification. This limitation originates from the inherent characteristics of reblurring supervision: smooth regions with minimal pixel-value variations may satisfy the optimization objective even in the absence of actual motion, potentially inducing spurious trajectory estimations (as empirically demonstrated in Fig.~\ref{fig:lim}(b)).
By providing additional auxiliary supervision, the integration of blur-aware mask constraints effectively alleviates this problem as shown in Fig.~\ref{fig:lim}(c). Our mask predictor introduces negligible parameters ($\textless 0.33$M), maintaining computational efficiency while resolving the motion estimation ambiguity. 
Comparative evaluations demonstrate that our approach achieves superior performance-efficiency trade-offs compared to existing deblurring methods. 
Future work will focus on improving the reblur loss formula and designing more reasonable constraints to eliminate the problem of flat areas being detected as blurred areas.

\section{More Visual Comparisons}
\begin{figure*} 
	\centering
	\begin{tabular}{c}
		\includegraphics[width=0.8\linewidth]{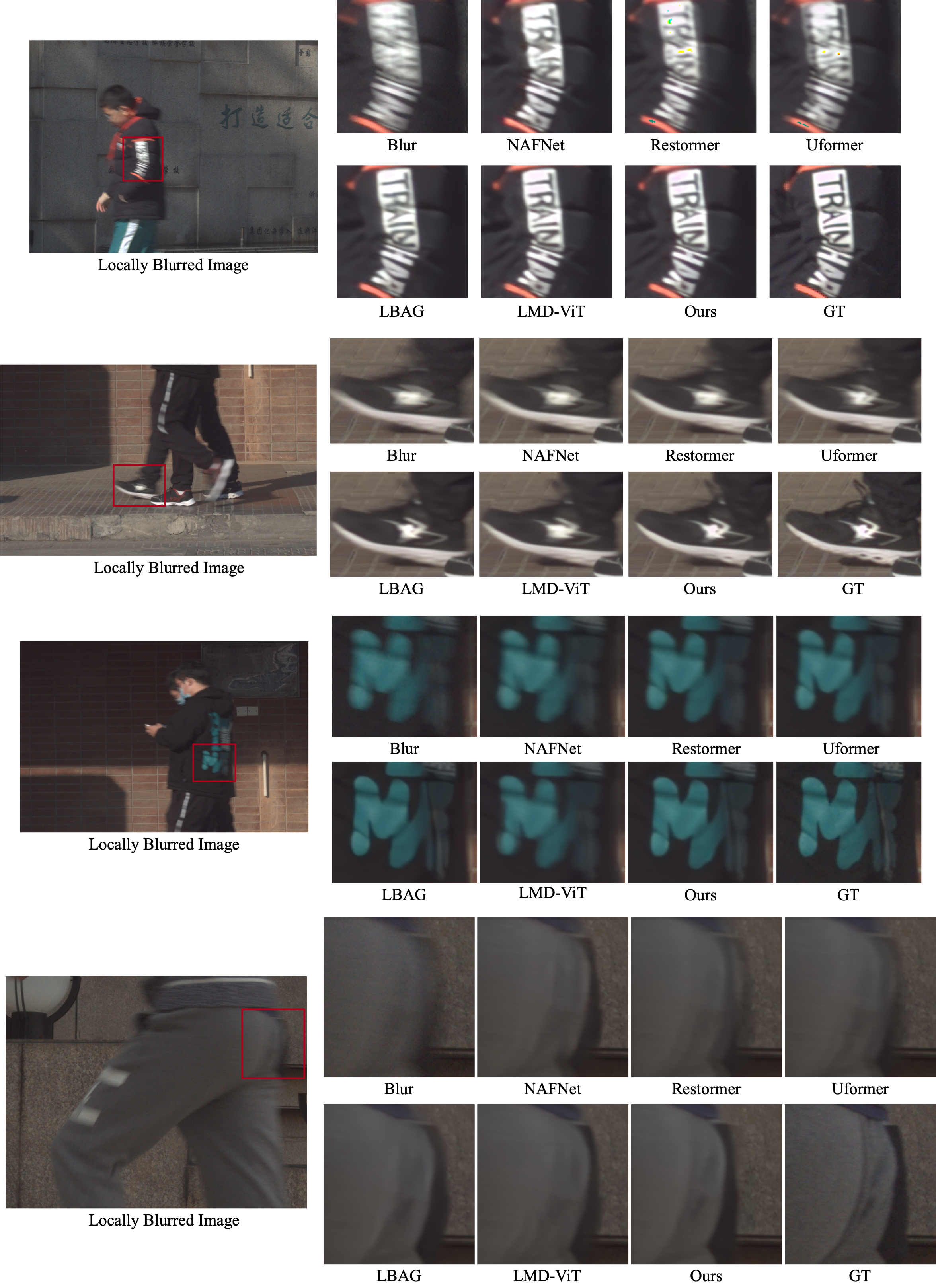}\\
	\end{tabular}
	\caption{Visual comparison with the state-of-the-art deblurring methods on ReLoBlur~\cite{li2023real}.
	}
	\label{fig:local_supp}
\end{figure*}

\begin{figure*} 
	\centering
	\begin{tabular}{c}
		\includegraphics[width=0.8\linewidth]{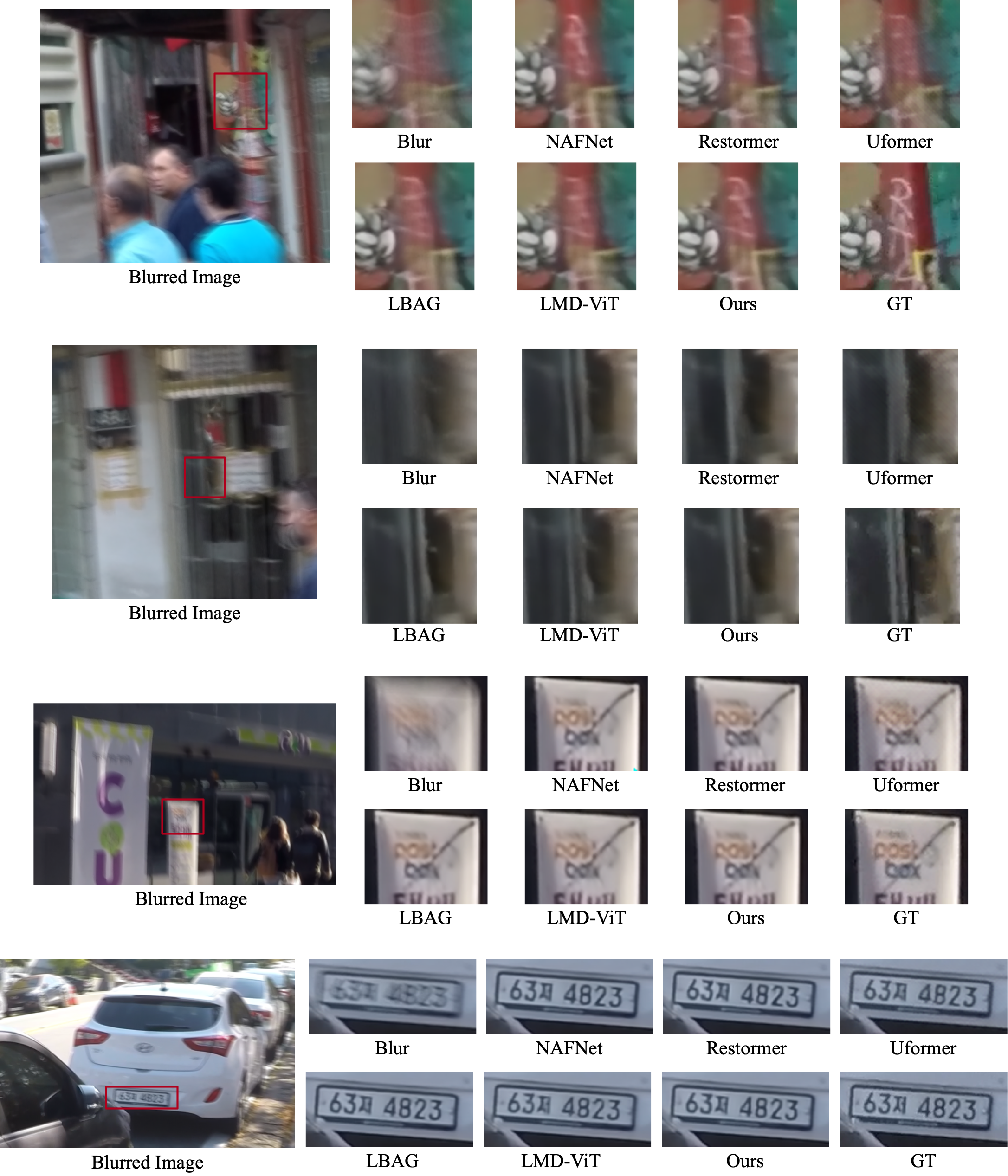}\\
	\end{tabular}
	\caption{Visual comparison with the state-of-the-art deblurring methods on GoPro~\cite{nah2017deep}. 
	}
	\label{fig:global_supp}
\end{figure*}

Our proposed M$^2$AENet addresses both local and global motion deblurring. Since blur masks and motion trajectories inherently indicate blurred regions and blur severity, our unified model adaptively handles varying motion blur patterns. We present extended comparisons of local and global deblurring results in Figs.~\ref{fig:local_supp} and~\ref{fig:global_supp}, respectively. 

As shown in Fig.~\ref{fig:local_supp}, our method precisely localizes blurred regions and restores details by leveraging estimated motion trajectories, achieving reconstruction quality closest to ground-truth (GT). Specific examples include recovered letters on clothing and brand logos on shoes, where traditional deblurring methods trained on mixed datasets fail to identify blur regions accurately. Their optimization objectives tend to converge to average solutions, resulting in poor detail recovery across most global deblurring approaches. Compared with existing local deblurring methods, our approach achieves comparable or superior visual quality while maintaining optimal performance-efficiency balance through adaptive pixel pruning, which minimizes computational costs.

Fig.~\ref{fig:global_supp} demonstrates that conventional global deblurring methods (e.g., NAFNet) achieve competitive performance in global restoration tasks. However, local deblurring methods like LMD-ViT exhibit limitations in global scenarios due to their window pruning strategies, which may incorrectly classify certain regions as sharp windows and skip processing. 

\begin{figure*} 
	\centering
	\begin{tabular}{c}
		\includegraphics[width=\linewidth]{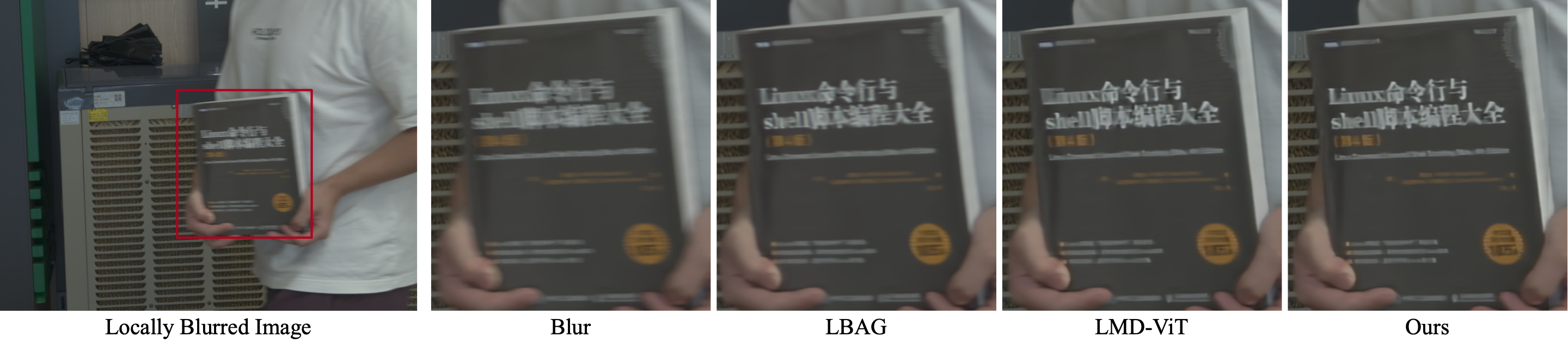}\\
	\end{tabular}
	\caption{Visual comparison with the state-of-the-art deblurring methods on real-world locally blurred images. 
	}
	\label{fig:real_supp}
\end{figure*}

\begin{figure*} 
	\centering
	\begin{tabular}{c}
		\includegraphics[width=0.8\linewidth]{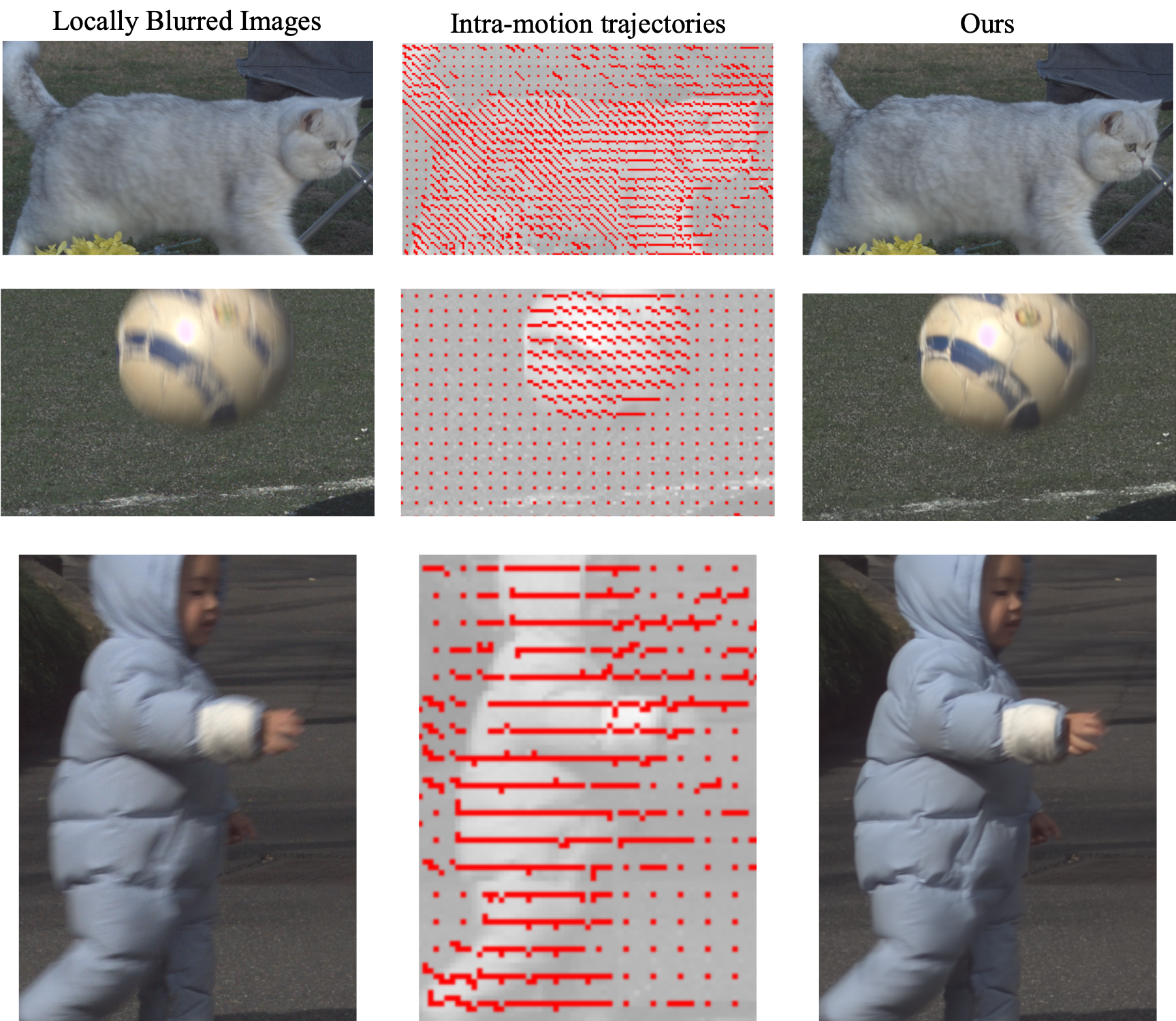}\\
	\end{tabular}
	\caption{Visualization of deblurring results and motion trajectories on ReLoBlur~\cite{li2023real}. 
	}
	\label{fig:offset_supp}
\end{figure*}

\begin{figure*} 
	\centering
	\begin{tabular}{c}
		\includegraphics[width=\linewidth]{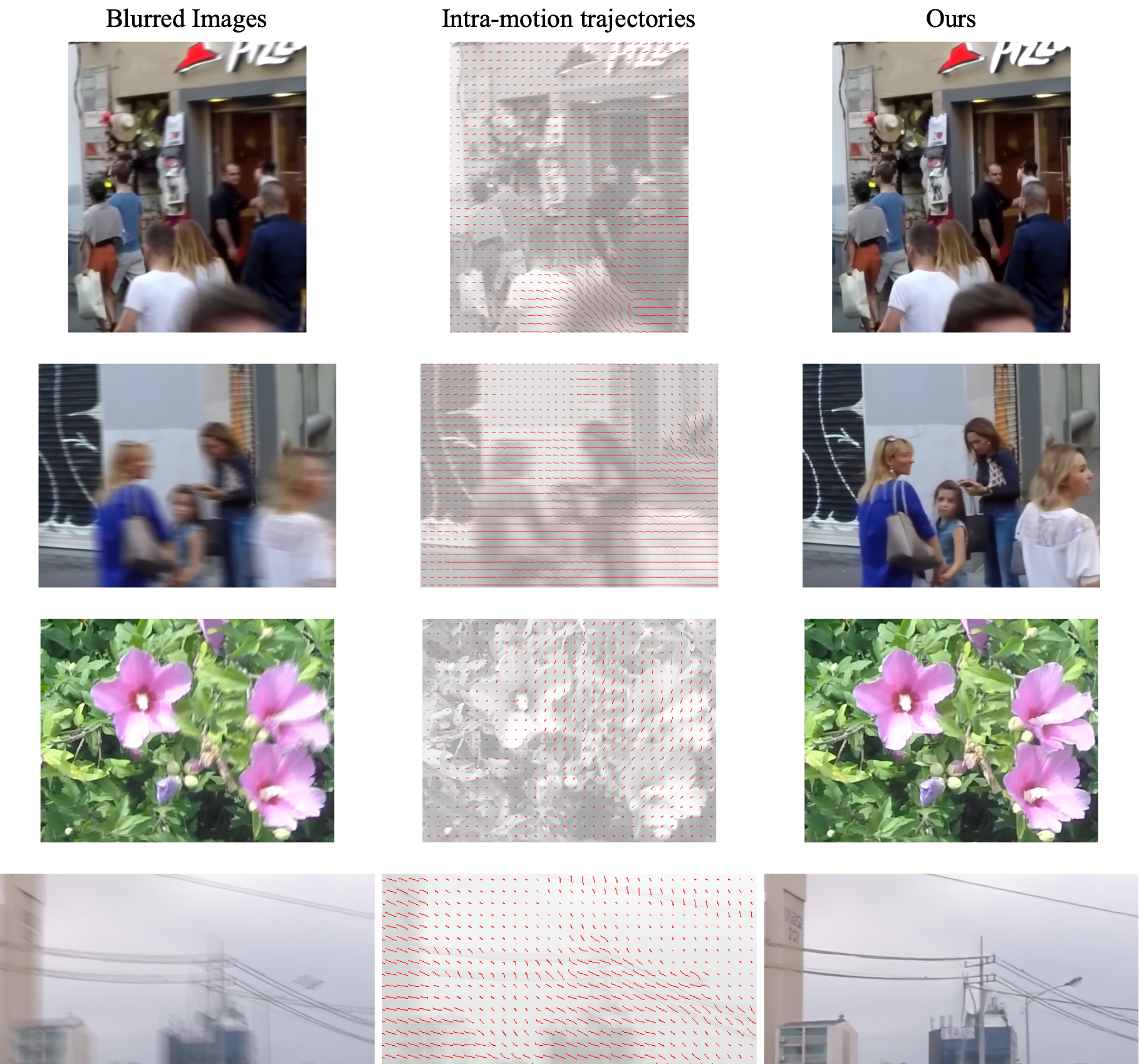}\\
	\end{tabular}
	\caption{Visualization of deblurring results and motion trajectories on GoPro~\cite{nah2017deep}. 
	}
	\label{fig:gopro_offset_supp}
\end{figure*}

Our method shows strong generalization on real-world smartphone images (Fig.~\ref{fig:real_supp}), successfully restoring book titles to legible states, outperforming existing local deblurring approaches. Additional visualizations in Figs.~\ref{fig:offset_supp} and ~\ref{fig:gopro_offset_supp} illustrate the capability of our method to locate blurred regions and analyze corresponding motion trajectories, such as a spinning soccer ball in mid-air and a cat walking across the grass. These results confirm the effectiveness of our approach in both motion trajectory estimation and blur removal.


\end{document}